# Identification of multi-scale hierarchical brain functional networks using deep matrix factorization

Hongming Li, Xiaofeng Zhu, Yong Fan

Center for Biomedical Image Computing and Analytics,
Department of Radiology, Perelman School of Medicine, University of Pennsylvania

**Abstract.** We present a deep semi-nonnegative matrix factorization method for identifying subject-specific functional networks (FNs) at multiple spatial scales with a hierarchical organization from resting state fMRI data. Our method is built upon a deep semi-nonnegative matrix factorization framework to jointly detect the FNs at multiple scales with a hierarchical organization, enhanced by group sparsity regularization that helps identify subject-specific FNs without loss of inter-subject comparability. The proposed method has been validated for predicting subject-specific functional activations based on functional connectivity measures of the hierarchical multi-scale FNs of the same subjects. Experimental results have demonstrated that our method could obtain subject-specific multi-scale hierarchical FNs and their functional connectivity measures across different scales could better predict subject-specific functional activations than those obtained by alternative techniques.

## 1 Introduction

The human brain can be represented as a multiscale hierarchical network [1, 2]. However, existing functional brain network analysis studies of resting-state functional magnetic resonance imaging (rsfMRI) data typically define network nodes at a specific scale, based on regions of interest (ROIs) obtained from anatomical atlases or functional brain parcellations [3-5]. Recent work has demonstrated important subject-specific variation in the functional neuroanatomy of large-scale brain networks [6], emphasizing the need for tools which can flexibly adapt to individuals' variation while simultaneously maintaining correspondence for group-level analyses.

To capture subject-specific FC without loss of inter-subject comparability, data-driven brain decomposition methods have been widely adopted to identify spatial intrinsic functional networks (FNs) and estimate functional network connectivity. In order to obtain subject-specific FNs from rsfMRI data of individual subjects while facilitating groupwise inference, independent vector analysis (IVA) and group-information guided ICA (GIGICA) methods have been proposed [7, 8]. More recently, several methods have been proposed to discover FNs from rsfMRI data with non-independence assumptions [9-11], and directly work on individual subject fMRI data and simultaneously enforce correspondence across FNs of different subjects by assuming that loadings of corresponding FNs of different subjects follow Gaussian [9] or delta-Gaussian [10] distributions. Non-negative matrix decomposition techniques have been adopted to simultaneously compute subject-specific FNs for a group of subjects regularized by group sparsity in order to separate anti-correlated FNs properly so that anti-correlation information between them could be preserved [11, 12]. However, these methods are not equipped to characterize multi-scale hierarchical organization of the brain networks [2]. Although clustering and module detection algorithms could be adopted to detect the hierarchical organization of brain networks, their performance is hinged on the network nodes/FNs used [13, 14].

To address the aforementioned limitations of existing techniques, we develop a novel brain decomposition model based on a collaborative sparse brain decomposition approach [11] and deep matrix factorization techniques [15], aiming to identify subject-specific, multi-scale hierarchal FNs from rsfMRI data. Based on rsfMRI data and task activation maps of unrelated subjects from the HCP dataset [16], we have quantitatively evaluated our method for predicting subject-specific functional activations based on functional connectivity measures of the hierarchical multi-scale FNs of the same subjects. Experimental results have demonstrated that the multi-scale hierarchical subject-specific FNs identified by our method from rsfMRI data could better predict the subject-specific functional activations evoked by different tasks than those identified by alternative techniques.

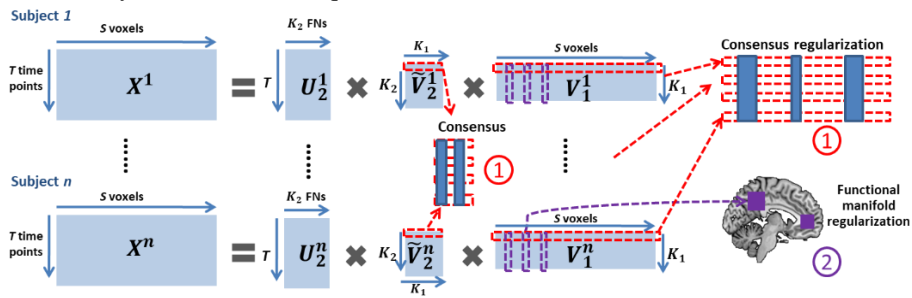

**Fig. 1.** Framework of the deep decomposition model for identifying multi-scale hierarchical subject-specific functional networks (a two-scale decomposition is illustrated).

## 2  Methods

A deep matrix decomposition framework is proposed to identify FNs at multiple scales as schematically illustrated in Fig. 1. Particularly, a deep semi-nonnegative matrix factorization is adopted to jointly detect a hierarchy of FNs from fine to coarse spatial scales in a data-driven way, and a group sparsity regularization term is adopted for FNs of different subjects at each scale to ensure the subject-specific FNs to share similar spatial patterns. Besides enforcing the groupwise correspondence of FNs across subjects, the group sparsity term also encourages FNs at the finest scale to have sparse spatial patterns and FNs at coarse scales to comprise functionally correlated FNs at finer scales. Our decomposition model is further enhanced by a data locality regularization term that makes the decomposition robust to imaging noise and improves spatial smoothness and functional coherences of the subject specific FNs.

### 2.1 Deep semi-nonnegative matrix factorization for brain decomposition

Given rsfMRI data $X^i \in R^{T \times S}$ of subject $i$, consisting of $S$ voxels and $T$ time points, we aim to find $K_j$ nonnegative FNs $V_j^i = (V_{j:k,s}^i) \in R_+^{K_j \times S}$ and their corresponding time courses $U_j^i = (U_{j:t,k}^i) \in R^{T \times K_j}$ at $j = 1, \ldots, h$ scales, so that $X^i \approx U_1^i V_1^i$; $X^i \approx U_2^i \tilde{V}_2^i V_1^i$, $V_2^i = \tilde{V}_2^i V_1^i$; $\cdots$; $X^i \approx U_h^i \tilde{V}_h^i \cdots V_1^i$, $V_h^i = \tilde{V}_h^i \cdots V_1^i$. The FNs $V_j^i$ at $1 \leq j \leq h$ scales are constrained to have a hierarchical structure and to be non-negative so that each FN does not contain any anti-correlated functional units. A deep semi-nonnegative matrix factorization (DSNMF) framework similar to [15] is adopted to identify the multi-scale hierarchical FNs by optimizing

$$\min_{\{U_j^i, \tilde{V}_j^i\}} \|X^i - U_h^i \tilde{V}_h^i \cdots \tilde{V}_1^i\|_F^2, \ s.t. \ \tilde{V}_j^i \geq 0, V_1^i = \tilde{V}_1^i, \forall 1 \leq j \leq h. \tag{1}$$

The FNs at different scales are represented by $V_h^i = \tilde{V}_h^i \cdots V_1^i, \forall 1 \leq j \leq h$ for subject $i$, and the FNs at 2 consecutive scales are linked hierarchically according to weights determined during the joint decomposition. The decomposition model does not constrain time courses $U_h^i$ to be non-negative so that it can be applied to preprocessed fMRI data with negative values.

### 2.2 DSNMF based collaborative brain decomposition

Given a group of $n$ subjects, each having fMRI data $X^i \in R^{T \times S}$, $i = 1, \ldots, n$, we identify subject-specific, multiscale hierarchical FNs by optimizing a joint model with integrated data fitting and regularization terms as illustrated in Fig. 1:

$$\min_{\{U_j^i, \tilde{V}_j^i\}} \sum_{i=1}^{n} \|X^i - U_h^i \tilde{V}_h^i \cdots \tilde{V}_1^i\|_F^2 + \sum_{j}^{h} \lambda_{c,j} R_{c,j} + \lambda_M \sum_{i=1}^{n} R_M^i, \tag{2}$$

$$s.t. \ \tilde{V}_j^i \geq 0, \left\|\tilde{V}_{j:k_j,\cdot}^i\right\|_\infty = 1, \forall 1 \leq k_j \leq K_j, \forall 1 \leq i \leq n, 1 \leq j \leq h,$$

where $\lambda_{c,j} = \alpha \cdot \frac{n \cdot T}{K_j}$ and $\lambda_M = \beta \cdot \frac{T}{K_1 \cdot n_M}$ are parameters for balancing data fitting and regularization terms, $T$ is the number of time points, $K_j$ is the number of FNs at scale $j$, $n_M$ is the number of spatially neighboring voxels at voxel level, $\alpha$ and $\beta$ are 2 parameters, $R_{c,j}$ and $R_M^i$ are regularization terms. Particularly, $R_{c,j}$ is an inter-subject group sparsity term that enforces FNs of different subjects to have common spatial structures at the same scale $j$. The group sparsity regularization term is defined as $R_{c,j} = \sum_{k=1}^{K_j} \left\|\widetilde{\tilde{V}_{j:k,\cdot}^{1..n}}\right\|_{2,1} = \sum_{k=1}^{K_j} \frac{\sum_{s=1}^{S_j}(\sum_{i=1}^{n}(\tilde{V}_{j:k,s}^i)^2)^{1/2}}{(\sum_{s=1}^{S_j} \sum_{i=1}^{n}(\tilde{V}_{j:k,s}^i)^2)^{1/2}}$ for each row of $\tilde{V}_j^i, \forall 1 \leq i \leq n, 1 \leq j \leq h$, where $S_j = K_{j-1}$ when $j > 1$ and $S_1 = S$. The group sparsity regularization (term 1 in Fig.1) enforces corresponding FNs of different subjects to have non-zero elements at the same spatial locations. Moreover, it encourages FNs to have spatially localized loadings. It is worth noting that the group sparsity term does not force different FNs to be non-overlapping, thus certain functional units may be included in multiple FNs simultaneously at each scale. We also adopt a data locality regularization term (term 2 in Fig.1), $R_M^i$, to encourage spatial smoothness and functional coherence of the FNs using a graph regularization technique [17] at the finest spatial scale, which is defined as $R_M^i = Tr(V_1^i L_M^i (V_1^i)')$, where $L_M^i = D_M^i - W_M^i$ is a Laplacian matrix for subject $i$, $W_M^i$ is a pairwise affinity matrix to measure spatial closeness or functional similarity between voxels, and $D_M^i$ is its corresponding degree matrix, the affinity between each pair of spatially connected voxels is calculated as $(1 + corr(X_{\cdot,a}^i, X_{\cdot,b}^i))/2$, where $corr(X_{\cdot,a}^i, X_{\cdot,b}^i)$ is the Pearson correlation coefficient between their rsfMRI signals, and others are set to zero so that $W_M^i$ had a sparse structure.

We optimize the joint model using an alternative update strategy. When $U_h^i$ and $\tilde{V}_k^i$ ($k \neq j$) are fixed, $\tilde{V}_j^i$ is updated as

$$\tilde{V}_j^i = \tilde{V}_j^i \odot \sqrt{\frac{[U_j^{i'} X^i]_+ \overline{V}_j^{i'} + [U_j^{i'} U_j^i]_- \tilde{V}_j^i \overline{V}_j^i \overline{V}_j^{i'} + \lambda_{c,j} \tilde{V}_j^i \odot G_j^{L21}/(G_j^{L2})^3}{[U_j^{i'} X^i]_- \overline{V}_j^{i'} + [U_j^{i'} U_j^i]_+ \tilde{V}_j^i \overline{V}_j^i \overline{V}_j^{i'} + \lambda_{c,j} \tilde{V}_j^i/(G_j \odot G_j^{L2})}} \; if \; j > 1$$

and $\tilde{V}_1^i = \tilde{V}_1^i \odot \sqrt{\frac{[U_1^{i'} X^i]_+ + [U_1^{i'} U_1^i]_- \tilde{V}_1^i + \lambda_M \tilde{V}_1^i W_M^i + \lambda_{c,1} \tilde{V}_1^i \odot G_1^{L21}/(G_1^{L2})^3}{[U_1^{i'} X^i]_- + [U_1^{i'} U_1^i]_+ \tilde{V}_1^i + \lambda_M \tilde{V}_1^i D_M^i + \lambda_{c,1} \tilde{V}_1^i/(G_1 \odot G_1^{L2})}} \; if \; j = 1$ (3)

$$G_{j:k,\cdot}^{L21} = repmat(\sum_{s=1}^S (\sum_{i=1}^n (\tilde{V}_{j:k,s}^i)^2)^{1/2}, 1, S),$$
$$G_{j:k,\cdot}^{L2} = repmat((\sum_{s=1}^S \sum_{i=1}^n (\tilde{V}_{j:k,s}^i)^2)^{1/2}, 1, S),$$
$$G_{j:k,s} = (\sum_{i=1}^n (\tilde{V}_{j:k,s}^i)^2)^{1/2}, \overline{V}_j^i = \tilde{V}_{j-1}^i \tilde{V}_{j-2}^i \cdots \tilde{V}_1^i,$$

where $\odot$ denotes element-wise multiplication, $repmat(b, r, c)$ denotes matrix obtained by replicating $r$ and $c$ copies of vector $b$ in the row and column dimensions, $[a]_+ = \frac{abs(a)+a}{2}$, $[a]_- = \frac{abs(a)-a}{2}$, $M_{j:k,\cdot}$ denotes the $k$-th row of matrix $M_j$, and $M_{j:k,s}$ denotes the $(k,s)$-th element in the matrix. $\tilde{V}_j^i$ is normalized by the row-wise maximum value along the row dimension after each update iteration.

When $\tilde{V}_k^i$, $1 \leq k \leq j$ are fixed, $U_j^i$ is updated as

$$U_j^i = X_j^i (\tilde{V}_j^i \tilde{V}_{j-1}^i \cdots \tilde{V}_1^i)^\dagger,$$ (4)

where $M^\dagger$ denotes the Moore-Penrose pseudo-inverse of matrix $M$.

To expedite the convergence of the optimization, a pre-training step at group level is adopted before the joint optimization. In particular, we compute $(U_1, V_1) \leftarrow sparseNMF(X)$ first, where $X$ denotes the temporal concatenated data of $\{X^i, i = 1, \ldots, n\}$, and $(U_2, \tilde{V}_2) \leftarrow sparseNMF(U_1), \cdots, (U_h, \tilde{V}_h) \leftarrow sparseNMF(U_{h-1})$, respectively. $\{\tilde{V}_1, \tilde{V}_2, \ldots, \tilde{V}_h\}$ are then used to initialize $\{\tilde{V}_j^i, i = 1, \ldots, n, j = 1, \ldots, h\}$ for the joint optimization. $\{\tilde{V}_1, \tilde{V}_2, \ldots, \tilde{V}_h\}$ contains a hierarchical structure with overlapping as they are obtained by decomposition in a greedy manner. Once the initialization is done, all FNs at different scales are optimized jointly. We set the parameters $\alpha$ to 1, and $\beta$ to 10 according to [11] in the present study.

## 3    Experimental results

We evaluated our method based on rsfMRI data and task activation maps of 40 unrelated subjects obtained from the Human Connectome Project (HCP) [16], aiming to evaluate the performance of predicting task-evoked activation responses based on functional connectivity measures of FNs at multiple scales. The FNs derived from rsfMRI data have demonstrated promising performance for predicting task-evoked brain activations [18].

The proposed decomposition model was applied to the minimal-preprocessed, cortical gray-coordinates based rsfMRI data. The number of scales was set to 3, and the number of FNs at the first scale was set to 90, which was estimated by MELODIC automatically [19], the numbers of FNs at the $2^{nd}$ and $3^{rd}$ scales were set to 50 and 25 respectively, which were decreased by half approximately along the scales.

We compared the proposed model with two alternative decomposition strategies: multi-scale decomposition performed independently at different scales, and greedy agglomerative hierarchical multi-scale decomposition, with the same setting of scales and the same numbers of FNs. For the independent decomposition, the collaborative sparse decomposition model was adopted to obtain subject-specific FNs independent at 3 scales (with independent initialization). For the agglomerative one, the initialization obtained by greedy decomposition as described section 2.2 was adopted for each scale, but the final decomposition at each scale was obtained using the collaborative sparse model separately.

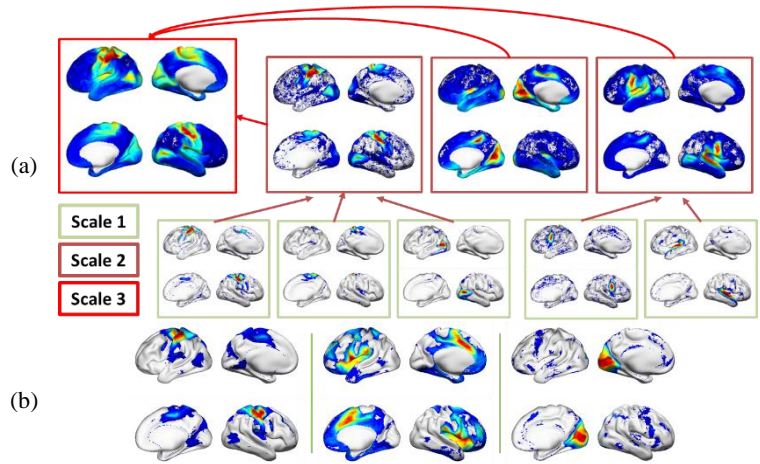

**Fig. 2.** Multi-scale functional networks identified by different methods. (a) A hierarchy of FNs obtained by the proposed method corresponding to sensorimotor networks. FNs at different scales are denoted by bounding boxes in different colors. (b) Sensorimotor regions in separate FNs at the coarsest scale (the $3^{rd}$) identified independently of other scales.

### 3.1 Multi-scale brain functional networks with a hierarchical organization

An example hierarchy of FNs (mean of 40 subjects) corresponding to sensorimotor function obtained by the proposed method is illustrated in Fig. 2 (a). The FN at the $3^{rd}$ scale (top left) comprised the sensorimotor networks and part of visual networks, and was a weighted composition of FNs at the $2^{nd}$ scale while the FNs at the $2^{nd}$ scale were composed of FNs at the $1^{st}$ scale. This example hierarchy of FNs illustrated that FNs corresponding to sensorimotor function gradually merged from fine to coarse scales in the hierarchy. However, no clear hierarchical organization was observed for

FNs independently identified at different scales. Particularly, as shown in Fig. 2 (b) sensorimotor regions appeared in separate FNs at the coarsest scale (the 3$^{rd}$), instead of forming a single FN. We postulated that the independent decomposition at different scales favored to better data-fitting and therefore was affected by data noise, while the joint decomposition model was more robust to data noise, facilitating accurate identification of FNs with coherent functions, such as the FN comprising sensorimotor regions shown in Fig. 2 (a).

### 3.2 Prediction of task evoked activations based on multi-scale FN connectivity

As no ground truth is available for FNs derived from rsfMRI data, we evaluated the multi-scale hierarchical FNs for predicting functional activations evoked by different tasks based on their functional connectivity measures with an assumption that better FNs could provide more discriminative information for predicting the brain activations. We also compared FNs obtained using different strategies in terms of their prediction performance.

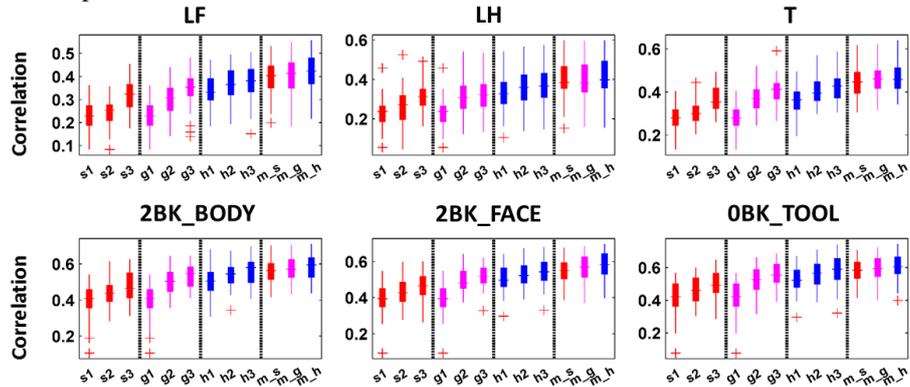

**Fig. 3.** Quantitative comparison of prediction performance for functional activations of different tasks (top: motor task events including left foot, left hand, and tongue; bottom: working memory task events including 2bk_body, 2bk_face, and 0bk_tool). s1 to s3, g1 to g3, and h1 to h3 denote models trained based on FNs at single scale identified independently, in a greedy agglomerative manner, and hierarchically by our method, respectively, while m_s, m_g and m_h denote models trained based on corresponding multi-scale FNs. The models built upon FNs obtained by our method performed significantly better than others (Wilcoxon signed rank test, $p < 0.02$).

Particularly, a whole brain voxelwise functional connectivity (FC) map was obtained for each FN by computing voxelwise Pearson correlation coefficient between the FN's time course and every cortical gray-coordinate's time course of the rsfMRI data. All the FC maps were then transferred to z-score maps using Fisher Z transformation. All the z-score values of FNs on each cortical gray-coordinate were used as features to predict its activation measures under different tasks of the same subject. Similar to [18], the whole cortical surface was divided into 90 parcels according to FNs obtained at the finest scale, and one ordinary least square model was trained for each parcel and every task event. The prediction performance was evaluated using a leave-one-subject-out cross-validation, where one subject's activation was predicted by a model built upon data of the remaining 39 subjects. The prediction was conducted using single-scale FNs (90, 50, 25) or multi-scale FNs (165) obtained using different strategies respectively, and the prediction accuracy was evaluated as the Pearson correlation coefficients between the predicted and real activation maps of 47 task events from 7 tasks categories.

The prediction performance of 6 randomly selected task events is illustrated in Fig. 3. For all task events and FNs identified by different strategies, the prediction models built upon multi-scale FNs outperformed all prediction models built upon any single scale FNs alone, indicating that multi-scale FNs could provide complementary information for the task activation prediction. The prediction models built upon multi-scale hierarchical FNs obtained by our method had significantly better performance than those build upon multi-scale FNs obtained by either the independent decomposition or the greedy agglomerative decomposition (Wilcoxon signed rank test, $p < 0.02$), indicating that the joint optimization of multi-scale hierarchical FNs could benefit from each other and characterize the intrinsic FNs better.

## 4      Conclusions

In this study, we have developed a deep decomposition model to identify multi-scale, hierarchical, subject-specific FNs with group level correspondence across different subjects. Our method is built upon deep semi-nonnegative matrix factorization framework, enhanced by a group sparsity regularization and graph regularization for maintaining inter-subject correspondence and better functional coherence. Experimental results based on rsfMRI data and task activation maps of the same subjects have demonstrated that the multi-scale hierarchical subject-specific FNs could capture informative intrinsic functional networks and improve the prediction performance of task activations evoked by different tasks, compared to FNs identified at different scales independently or in a greedy agglomerative way. In conclusion, our method provides an improved solution for characterizing subject-specific, multi-scale hierarchical organization of the brain functional networks.

## 5      Acknowledgements

This work was supported in part by National Institutes of Health grants [CA223358, EB022573, DK114786, DA039215, and DA039002].